\documentclass{article}

\usepackage{iclr2027_conference,times}

\iclrfinalcopy

\usepackage{amsmath,amsfonts,bm}

\def\eqref#1{equation~\ref{#1}}

\def\1{\bm{1}}

\DeclareMathAlphabet{\mathsfit}{\encodingdefault}{\sfdefault}{m}{sl}
\SetMathAlphabet{\mathsfit}{bold}{\encodingdefault}{\sfdefault}{bx}{n}

\usepackage{amssymb}
\usepackage{hyperref}
\hypersetup{hidelinks,pdftitle={VideoTapestry: Query-Adaptive Memory Refinement for Multi-Agent Long-Video Understanding},pdfauthor={Yucheng Liu, Yufei Yin, Mingxiao Feng, Jiajun Deng, Wengang Zhou, Houqiang Li}}
\usepackage{url}
\usepackage{graphicx}
\usepackage{xcolor}
\usepackage{capt-of}
\usepackage{booktabs}
\usepackage{array}
\usepackage[most]{tcolorbox}
\usepackage{listings}
\usepackage{textcomp}

\newlength{\lvbenchpendingwidth}
\definecolor{gaincolor}{RGB}{31,132,82}
\definecolor{groupbeige}{RGB}{239,236,232}
\definecolor{grouplilac}{RGB}{236,234,238}
\definecolor{groupgreen}{RGB}{232,238,233}
\definecolor{groupblue}{RGB}{228,234,241}
\definecolor{oursblue}{RGB}{219,228,239}
\definecolor{overviewblue}{RGB}{0,98,133}
\definecolor{skimgreen}{RGB}{21,127,5}
\definecolor{focusbrown}{RGB}{208,69,0}
\definecolor{sharedpurple}{RGB}{105,72,150}
\definecolor{answergray}{RGB}{78,82,88}
\newcommand{\scoregain}[2]{\textbf{#1}\,\textcolor{gaincolor}{\raisebox{0.08ex}{\scalebox{0.68}{(+\textbf{#2})}}}}

\newtcblisting{agentpromptbox}[2]{
  enhanced,
  listing only,
  breakable,
  title={#1},
  colback=#2!3!white,
  colframe=#2,
  colbacktitle=#2!14!white,
  coltitle=#2,
  fonttitle=\bfseries,
  arc=2.5mm,
  outer arc=2.5mm,
  boxrule=0.8pt,
  titlerule=0.8pt,
  left=2.5mm,
  right=2.5mm,
  top=1.8mm,
  bottom=1.8mm,
  listing options={
    basicstyle=\ttfamily\footnotesize,
    breaklines=true,
    columns=fullflexible,
    keepspaces=true,
    upquote=true,
    escapeinside={(*@}{@*)},
    literate={"}{{{\fontencoding{T1}\selectfont\textquotedbl}}}1,
    showstringspaces=false
  }
}

\makeatletter
\newcommand{\tableband}[1]{\noalign{\begingroup\color{#1}%
  \hrule height\dimexpr\ht\@arstrutbox+\dp\@arstrutbox\relax
  \kern-\dimexpr\ht\@arstrutbox+\dp\@arstrutbox\relax\endgroup}}
\makeatother

\title{%
VideoTapestry: Query-Adaptive Memory  \\Refinement for Multi-Agent Long-Video \\Understanding
}

\author{%
  \textbf{Yucheng Liu}\textsuperscript{1, 2},
  \textbf{Yufei Yin}\textsuperscript{3},
  \textbf{Mingxiao Feng}\textsuperscript{4},
  \textbf{Jiajun Deng}\textsuperscript{1, 4, \dag}, \\
  \textbf{Wengang Zhou}\textsuperscript{1, 4},
  \textbf{Houqiang Li}\textsuperscript{1, 4}\\[0.8em]
  {\small\textsuperscript{1}University of Science and Technology of China}\\
   {\small\textsuperscript{2}Jilin University}\\
  {\small\textsuperscript{3}Hangzhou Dianzi University}\\
  {\small\textsuperscript{4}Institute of Artificial Intelligence, Hefei Comprehensive National Science Center}%
}

\begin{document}

\maketitle
\lhead{}

\begingroup
\renewcommand{\thefootnote}{}
\footnotetext{%
  \parbox[t]{\dimexpr\linewidth-0.3em\relax}{%
    This work was performed when Yucheng Liu was a research intern at USTC.\par
    \dag\,\, Corresponding author%
  }%
}
\endgroup

\begin{abstract}

Long-video understanding places substantial demands on memory, as answering questions often requires retrieving information distributed across extended temporal spans. Existing approaches broadly follow two paradigms: query-driven exploration, which is sensitive to localization errors, and query-independent memory construction, which may omit question-specific details.
We introduce \textbf{VideoTapestry}, a training-free multi-agent framework that adapts a preconstructed hierarchical video memory through coarse-to-fine, query-driven refinement. 
The preconstructed memory organizes video content into three levels, capturing global narrative context, event-level temporal structure, and fine-grained relational evidence, respectively.
To support coarse-to-fine localization and observation, we assign a specialized agent to each level, keeping retrieval and refinement within a scale-specific context. Guided by the query, these agents revisit relevant video regions and enrich layer-wise memories with targeted multimodal observations. Their refinements are assembled according to the original hierarchy into a composite query-adaptive memory, preserving global context in a compact form while retaining fine-grained evidence along query-relevant branches for final reasoning.
Compared with direct GPT-5.5 inference, VideoTapestry achieves absolute accuracy gains of 17.2\%, 14.9\%, 9.8\%, and 7.0\% on LVBench, LongVideoBench (Long), Video-MME (Long), and EgoSchema, respectively, achieving the state-of-the-art results among all competitors. 

\end{abstract}

\section{Introduction}
\label{sec:introduction}

Long-video understanding is a fundamental capability for vision-language models (VLMs) reasoning over extended activities, narratives, and real-world interactions~\citep{fu2025video,wu2024longvideobench,wang2025lvbench,mangalam2023egoschema}. Compared with short clips, answering questions about hour-long videos requires preserving global context while locating decisive evidence at appropriate semantic granularity. Uniform sampling may miss brief events, while dense frame sampling introduces redundancy and query-irrelevant information, making relevant content difficult to isolate~\citep{ye2025re,wang2025videotree,wu2024longvideobench}. These challenges motivate memory mechanisms that organize video information for selective retrieval while preserving context for reasoning.

\begin{figure}[t]
\centering
\includegraphics[width=\textwidth]{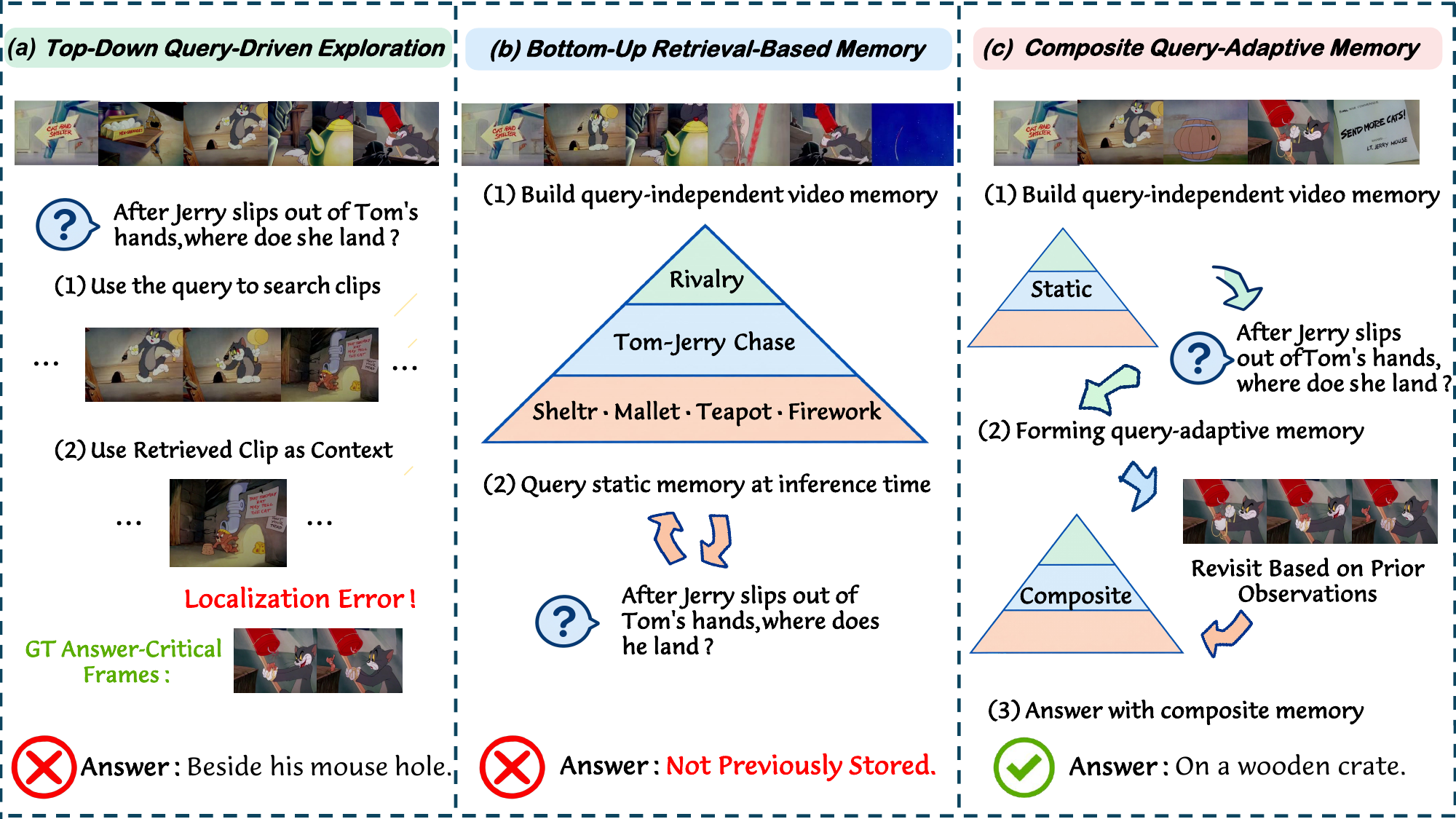}
\caption{\textbf{Comparison of long-video context construction paradigms.}
\textbf{Left:} Top-down exploration may miss answer-critical evidence due to localization errors.
\textbf{Middle:} Bottom-up memory provides broad coverage but may omit query-specific details.
\textbf{Right:} VideoTapestry combines a query-independent hierarchy with targeted revisiting to form a composite query-adaptive memory that preserves global context and enriches relevant evidence.}
\label{fig:paradigm-comparison}
\end{figure}

Existing approaches broadly fall into two paradigms. As shown in Figure~\ref{fig:paradigm-comparison}(a), top-down methods start from the query and inspect relevant intervals, constructing or updating hierarchical memory during search~\citep{wang2025videotree,ye2025re,yin2026videoarm,lin2026videoseek}.  
This enables adaptive observation, but inaccurate localization may direct the search away from key video segments.
Bottom-up methods, as illustrated in Figure~\ref{fig:paradigm-comparison}(b), construct a query-independent video memory and retrieve or revisit relevant content at inference time~\citep{an2024videoagent,zuo2026videolucy,zhang2026deep,yeo2026worldmm}.
Such representations provide broad coverage but may omit details required by specific questions. Intuitively, the paradigms are complementary: query-independent memory provides broad coverage, while query-driven exploration acquires question-specific evidence.


However, integrating query-specific evidence with existing memory while preserving global context remains challenging. Integrating the strengths of the two paradigms requires incorporating new evidence into memory while limiting irrelevant content. This process involves global localization, temporal narrowing, and fine-grained verification. These complementary reasoning objectives motivate specialized agents that refine memory at corresponding semantic levels and pass inspection objectives between levels.

Motivated by these considerations, we introduce \textbf{VideoTapestry}, a training-free multi-agent framework that keeps a preconstructed hierarchy fixed while transforming it into a composite query-adaptive memory through coarse-to-fine, query-driven refinement.
The central problem we address is how to adapt a query-independent hierarchical memory to a specific question without sacrificing its global coverage, while selectively acquiring and integrating fine-grained evidence that is missing or insufficient in the original memory.

The paradigm of VideoTapestry is illustrated in
Figure~\ref{fig:paradigm-comparison}(c). The memory consists of
SuperEvents, MacroEvents, and local Subgraphs, representing global
stages, localized events, and fine-grained relational evidence.
The Overview, Skim, and Focus agents operate at these levels,
respectively, each within a scale-specific context. Guided by candidate
memories and upstream inspection objectives, each agent acquires
targeted multimodal evidence, refines a temporary layer memory, and
decides whether to stop or pass selected branches and unresolved
inspection objectives to the next level.

At the Super level, relevant descriptions and their neighboring context
are retained and enriched, while the remaining events are summarized
to preserve global coverage. Along selected branches, MacroEvent and
Subgraph descriptions are refined with question-specific evidence.
These layer-wise memories are assembled according to their original
parent-child relations and paired with a concise query-time trace of
inspection objectives and stopping decisions. The resulting
\textbf{composite query-adaptive memory} combines a refined global
scaffold with selectively refined local paths, preserving global
context while retaining detailed query-relevant evidence. All
refinements are query-specific and leave the preconstructed memory
unchanged. An independent Answer Agent generates the final prediction
from the assembled memory without further retrieval or observation.

We evaluate VideoTapestry on LVBench, LongVideoBench, Video-MME, and EgoSchema against representative direct VLM baselines and agentic approaches. Using GPT-5.5 as the base VLM, VideoTapestry improves accuracy over direct GPT-5.5 inference by 17.2, 14.9, 9.8, and 7.0 percentage points on LVBench, LongVideoBench (Long), Video-MME (Long), and EgoSchema, respectively. It achieves the best reported results among the compared methods across all four benchmarks. Controlled ablation studies further demonstrate the effectiveness of query-adaptive memory refinement and scale-aligned multi-agent collaboration.

Our contributions are threefold:

\begin{itemize}
\item We propose a query-adaptive memory strategy that integrates newly observed evidence into selected branches while preserving global context and the original hierarchical structure.

\item We develop \textbf{VideoTapestry}, a training-free multi-agent framework that aligns agents with memory levels and coordinates refinement through branch selection and inspection-objective transfer.

\item We demonstrate consistent improvements across four video understanding benchmarks, with controlled ablations and cross-backbone evaluations supporting the effectiveness of memory refinement and scale-aligned collaboration.
\end{itemize}

\section{Related Work}

\subsection{Long-Video Understanding}

Long-video understanding extends the scope of VLM reasoning from short clips to complex real-world streams lasting many minutes or even hours. Recent work expands visual context windows, compresses redundant spatiotemporal tokens, and scales long-form instruction tuning through LongVU~\citep{shen2024longvu}, LongViTU~\citep{wu2025longvitu}, and Hour-LLaVA~\citep{lin2026unleashing}. Video Panels and Tempo further improve effective temporal coverage via frame grids and query-aware visual compression~\citep{doorenbos2026video,fei2026small}. Despite these advances, practical context budgets still limit reliable evidence coverage, motivating agentic and memory-based methods that organize video content externally and acquire evidence through query-conditioned retrieval and observation~\citep{an2024videoagent,zhang2026deep,chen2026memdreamer}.

\subsection{Hierarchical Video Memory}

Video memory systems store frame descriptions, clips, or temporal segments outside the model context, while hierarchical and graph-based variants organize them across multiple semantic levels and explicit relations~\citep{long2026seeing,yeo2026worldmm,wang2025videotree,zuo2026videolucy,yin2026videoarm}. PyraVid uses structure-guided expansion and pruning~\citep{yan2026pyravid}, while VAM aligns temporal and spatial evidence~\citep{li2026visual}. MAGIC-Video combines multimodal graph retrieval with narrative chains~\citep{li2026bridging}, while MERIT uses multi-key episodic retrieval with selective query-time filtering~\citep{choi2026keep}. VideoTapestry adapts a query-independent hierarchy through targeted re-observation, retaining a compact global scaffold and refined query-relevant branches in a composite query-adaptive memory.

\subsection{Agentic and Multi-Agent Video Reasoning}

Agentic video systems replace one-pass prediction with planning, temporal search, retrieval, and selective observation~\citep{an2024videoagent,ye2025re,zhang2026deep,lin2026videoseek}. Active Video Perception plans observations and checks evidence sufficiency~\citep{wang2026active}, while LensWalk controls temporal scope and sampling density~\citep{li2026lenswalk}. Route2Look routes queries among global browsing, temporal grounding, and semantic retrieval via a routing skill~\citep{wang2026routing}. VideoXAgent acquires multimodal evidence online through expert tools~\citep{yang2026online}. Multi-agent approaches distribute perception, localization, verification, and reasoning across specialized roles~\citep{yan2026symphony,rege2026agentic,zuo2026videolucy}. VideoTapestry focuses on final hierarchical representation: scale-aligned agents retain a global scaffold, refine query-relevant branches, and assemble layer-wise memories for an independent Answer Agent.


\begin{figure}[t]
\centering
\includegraphics[width=\textwidth]{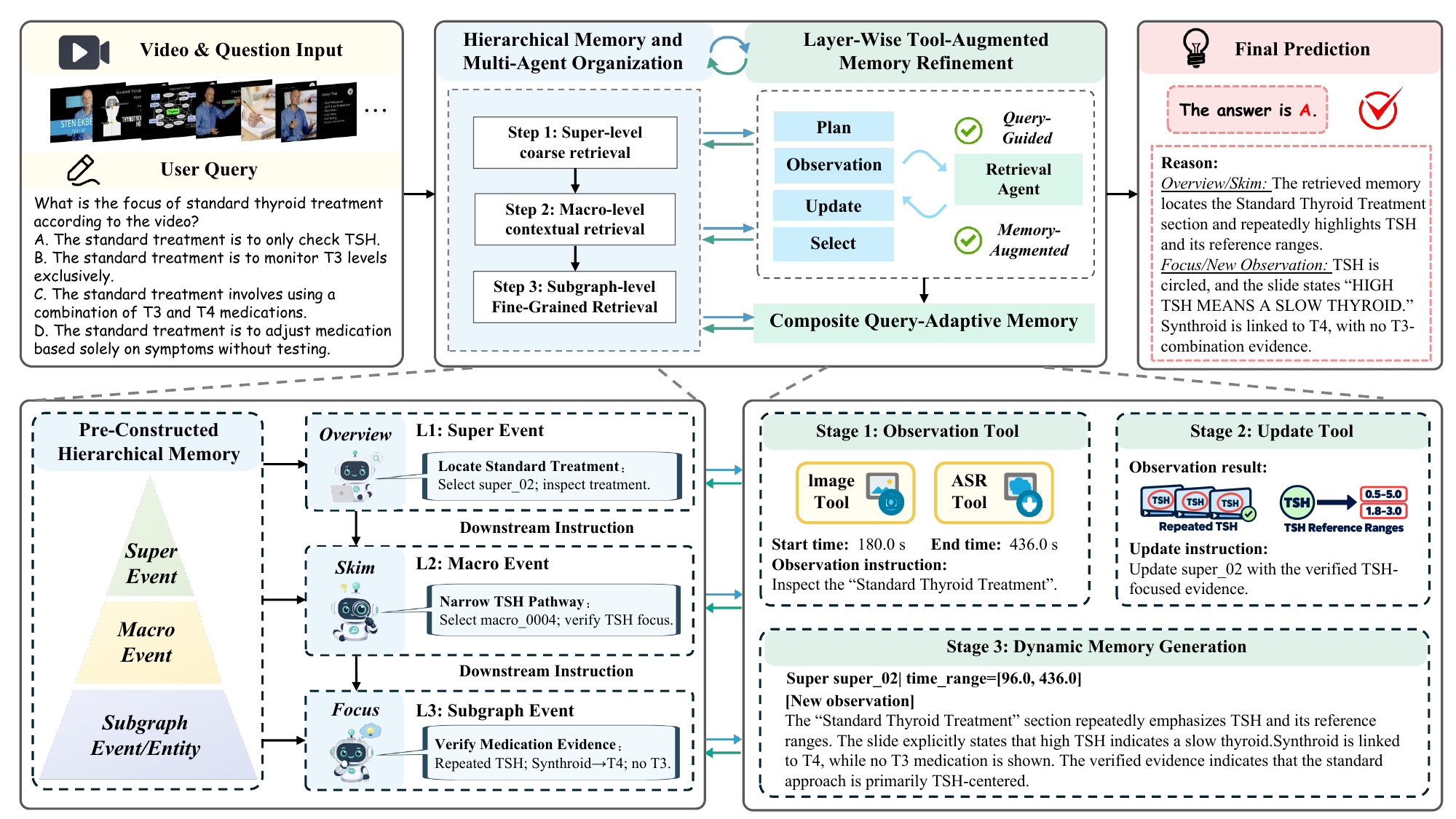}
\caption{\textbf{Overview of VideoTapestry.} Scale-aligned agents operate over a preconstructed hierarchical video memory, while layer-wise tool-augmented refinement selectively incorporates newly acquired query-relevant observations at different semantic levels. The resulting refined layer memories are hierarchically assembled into a composite query-adaptive memory for final prediction.}
\label{fig:framework-overview}
\end{figure}

\section{Method}

Given a long video \(V\), a question \(q\), and answer options \(o\),
VideoTapestry transforms a preconstructed hierarchical video memory
into a composite query-adaptive memory. Three scale-aligned agents
acquire evidence and refine temporary layer memories, which are then
assembled for an independent Answer Agent
(Figure~\ref{fig:framework-overview}).

\subsection{Query-Adaptive Memory and Scale-Aligned Organization}

\paragraph{Query-adaptive memory representation.}
VideoTapestry constructs a query-adaptive reasoning context by
compressing unrelated SuperEvents, enriching relevant descriptions
with new observations, and assembling refined descendants under
their original parents. These operations modify only temporary textual
representations while preserving the underlying static hierarchy.

We instantiate a fixed query-independent hierarchy following the
memory-construction procedure of MemDreamer~\citep{chen2026memdreamer},
upon which VideoTapestry performs query-adaptive refinement.
The hierarchy
\(\mathcal{G}=(\mathcal{V},\mathcal{E})\) contains three semantic levels:
\[
\mathcal{M}_{\mathrm{static}}=
\{\mathcal{M}_{\mathrm{static}}^{S},
\mathcal{M}_{\mathrm{static}}^{M},
\mathcal{M}_{\mathrm{static}}^{G}\}.
\]
SuperEvents summarize coarse temporal stages, MacroEvents describe
localized coherent events, and each MacroEvent is associated with
a complete local Subgraph containing Entity, Event, and OCRText
nodes connected by typed relations. The levels \(S\), \(M\), and
\(G\) provide global narrative context, event-level temporal
structure, and fine-grained factual evidence, respectively.

\paragraph{Scale-aligned multi-agent organization.}
The Overview Agent refines the Super-level scaffold and selects
relevant stages; the Skim Agent examines their MacroEvents and
narrows candidate regions; and the Focus Agent inspects the
corresponding Subgraphs for fine-grained verification. Each agent
operates within a scale-specific context, with selected parent
identifiers determining which descendants are exposed at the next level.

At level \(\ell\in\{S,M,G\}\), the agent produces a temporary
refined memory, selected identifiers, a downstream instruction,
and a stop flag:
\[
\left(
\widetilde{\mathcal{M}}_{q}^{\ell},
\mathcal{S}^{\ell},
\mathcal{I}^{\ell},
z^{\ell}
\right)
=
\mathcal{A}_{\ell}\!\left(
q,o,s_V,
\mathcal{M}_{\mathrm{static}}^{\ell}[\mathcal{C}^{\ell}],
\mathcal{I}^{\ell-1},
\mathcal{T}_{\ell}
\right).
\]
Here, \(s_V\), \(\mathcal{C}^{\ell}\),
\(\mathcal{I}^{\ell-1}\), and \(\mathcal{T}_{\ell}\) denote
the global video summary, candidate nodes, upstream instruction,
and level-specific tools, respectively. If refinement continues,
the selected identifiers define
\(\mathcal{C}^{\ell+1}=\bigcup_{v\in\mathcal{S}^{\ell}}
\operatorname{Children}(v,\mathcal{G})\),
which the next agent loads from static memory.
The instruction \(\mathcal{I}^{\ell}\) conveys the inspection
objective and may summarize intermediate findings and unresolved
evidence needs without transmitting the full preceding-layer
memory or tool trajectory.

The validated stop flag \(z^{\ell}\) determines whether refinement
terminates. In the evaluated Video-MME configuration, global questions
may terminate at an intermediate level, whereas non-global questions
must descend when valid descendants are available. The Focus Agent
always terminates exploration. The question-type heuristic and decision
validation rules are detailed in
Appendix~\ref{app:inference_details}.

\subsection{Layer-Wise Tool-Augmented Memory Refinement}

All agents follow the
\(\operatorname{Plan}_{\ell}\!\rightarrow\!
\operatorname{Observe}_{\ell}\!\rightarrow\!
\operatorname{Update}_{\ell}\!\rightarrow\!
\operatorname{Decide}_{\ell}\)
protocol. Plan identifies evidence needs, Observe inspects selected
intervals, Update integrates new evidence into temporary memory, and
Decide selects the next scope or terminates. Across all benchmarks, each visited layer performs one refinement batch, followed by one layer-level decision
(Appendix~\ref{app:inference_details}).

All levels access the source video through a Visual Observation Tool:
\[
\mathcal{T}_{\mathrm{vis}}^{\ell}
\left(
\mathcal{W}_{\mathrm{vis}}^{\ell},
\mathcal{I}_{\mathrm{obs}}^{\ell}
\right)
\rightarrow
\mathcal{O}_{q}^{\mathrm{vis},\ell},
\qquad \ell\in\{S,M,G\},
\]
where \(\mathcal{W}_{\mathrm{vis}}^{\ell}\) denotes a set of temporal
intervals, \(\mathcal{I}_{\mathrm{obs}}^{\ell}\) specifies what to
inspect, and \(\mathcal{O}_{q}^{\mathrm{vis},\ell}\) contains the
resulting observations.

The Overview Agent additionally accesses a Transcript/ASR Tool:
\[
\mathcal{T}_{\mathrm{txt}}
\left(
\mathcal{W}_{\mathrm{txt}},
\mathcal{I}_{\mathrm{txt}}
\right)
\rightarrow
\mathcal{O}_{q}^{\mathrm{txt}},
\]
which extracts spoken-information evidence from timestamped text
within selected windows. In the evaluated Video-MME configuration,
it uses Whisper transcripts and follows visual observation on each
successful Overview pass. The Skim and Focus Agents use visual
observations only.

\subsubsection{Overview Agent: Super-Level Memory Refinement}

The Overview Agent accesses all SuperEvents. For local questions,
it decomposes the localization objective into atomic targets.
Entity, OCR-text, and semantic retrieval interfaces provided by the
preconstructed memory~\citep{chen2026memdreamer} yield temporal
anchors for observation; retrieval outputs serve as localization
hints rather than verified answer evidence.

A visual planner converts these anchors into an inspection instruction
and a continuous observation window bounded by the video duration.
Sampled frames are arranged into grids for
\(\mathcal{T}_{\mathrm{vis}}^{S}\). In the evaluated Video-MME
configuration, a subsequent ASR planner uses the SuperEvent timeline
and visual results to select transcript windows and formulate a
spoken-information request. The resulting observations are
\[
\mathcal{O}_{q}^{S}
=
\mathcal{O}_{q}^{\mathrm{vis},S}
\cup
\mathcal{O}_{q}^{\mathrm{txt}}.
\]

Observations are associated with SuperEvents using candidate
identifiers and temporal overlap. Let \(\mathcal{C}_{\mathrm{ev}}\)
index the evidence-associated SuperEvents in temporal order.
A protected set includes these events and their immediate neighbors:
\[
\mathcal{P}
=
\left(
\bigcup_{i\in\mathcal{C}_{\mathrm{ev}}}
\{i-1,i,i+1\}
\right)
\cap \{1,\ldots,N_S\},
\]
where \(N_S\) is the number of SuperEvents.

The Overview Agent updates the Super-level memory asymmetrically.
Protected nodes retain their original descriptions and receive
available associated evidence, while the remaining nodes are compressed:
\[
\widetilde{m}_{i}^{S}
=
\begin{cases}
m_{i}^{S}\oplus
[\mathrm{Visual}]\,v_i\oplus
[\mathrm{Transcript}]\,a_i\oplus
[\mathrm{Evidence}]\,e_i,
& i\in\mathcal{P},\\[2mm]
c(m_i^S),
& i\notin\mathcal{P}.
\end{cases}
\]
Here, \(m_i^S\) is the original description, \(v_i\) contains visual
observations, \(a_i\) contains transcript-derived facts, and \(e_i\)
contains timestamped verbatim transcript evidence.
The operator \(\oplus\) denotes textual composition, and \(c(\cdot)\)
produces a compact summary. Unavailable evidence fields are omitted.
The protected set is enforced programmatically: protected neighbors
without associated observations retain their original text.

This refined global scaffold combines video-wide coverage with
query-relevant detail. The agent then selects SuperEvents and
a downstream instruction, or terminates according to the validated
stopping rule.

\subsubsection{Skim Agent: Macro-Level Memory Refinement}

The Skim Agent operates on
\(\mathcal{C}^{M}=
\bigcup_{v\in\mathcal{S}^{S}}\operatorname{Children}(v,\mathcal{G})\),
the MacroEvent children of the selected SuperEvents. Guided by the upstream instruction, it plans observations within
candidate intervals and associates the resulting evidence with the
covered nodes. For each requested update, it rewrites the original
description:
\[
\widetilde{\operatorname{text}}_{i}^{M}
=
\operatorname{Rewrite}_{M}\!\left(
\operatorname{text}_{i}^{M},
q,o,\mathcal{I}^{S},\mathcal{O}_{i}^{M}
\right).
\]
The rewrite preserves relevant static facts, incorporates observed
evidence, and compresses unrelated content. Missing or invalid updates
retain the original text. Node identifiers, parent relations, semantic
levels, and temporal ranges remain fixed. The agent then selects
MacroEvents and generates an instruction for finer inspection, subject
to the stopping rule.

\subsubsection{Focus Agent: Subgraph-Level Memory Refinement}

The Focus Agent loads local Subgraphs associated with the
selected MacroEvents, exposing entities, events, OCR content, and
typed relations. Guided by the upstream instruction, it plans
fine-grained visual inspection and rewrites the serialized description
of each requested Subgraph:
\[
\widetilde{\operatorname{text}}_{i}^{G}
=
\operatorname{Rewrite}_{G}\!\left(
\operatorname{text}_{i}^{G},
q,o,\mathcal{I}^{M},\mathcal{O}_{i}^{G}
\right).
\]
This operation modifies the temporary textual description rather than
the underlying graph nodes or edges. As at the Macro level, missing or
invalid updates retain the original text. The Focus Agent then
terminates hierarchical exploration.

\subsection{Composite Memory Assembly and Answer Generation}

The key representation in VideoTapestry is a composite query-adaptive
memory combining a refined global scaffold with selectively refined
local paths. After traversal, \(\operatorname{Assemble}\) retains the
refined Super-level scaffold and places selected refined
MacroEvent and Subgraph memories beneath their original parents,
preserving video-wide coverage while enriching query-relevant branches.

The resulting memory is serialized with the question, options, and a
compact handoff trace:
\[
\mathcal{D}_{q}
=
\operatorname{Serialize}\!\left(
q,o,H_q,
\operatorname{Assemble}\!\left(
\left\{
\widetilde{\mathcal{M}}_{q}^{\ell}
\right\}_{\ell\in\mathcal{L}_{q}}
\right)
\right).
\]
Here, \(\mathcal{L}_{q}\subseteq\{S,M,G\}\) denotes the visited levels,
and \(H_q=[(a_{\ell},i_{\ell})]_{\ell\in\mathcal{L}_{q}}\) records each
agent and its handoff instruction or stopping marker. The trace may
summarize intermediate findings and unresolved evidence needs, but
excludes complete observation and update logs.

The independent Answer Agent predicts
\[
\widehat{y}
=
\mathcal{A}_{\mathrm{ans}}(q,o,\mathcal{D}_{q})
\]
without further retrieval or observation.

\noindent\begin{minipage}{\textwidth}
\centering
\captionof{table}{Comparison with representative VLMs and agentic approaches on four long-video understanding benchmarks. Results are reported as accuracy (\%). Green parenthesized values denote percentage-point gains over GPT-5.5 (Direct). \(\dagger\) denotes our same-backbone end-to-end evaluation with frozen GPT-5.5. \(\ddagger\) denotes results from our reproduction. w/o: no provided subtitles.}
\label{tab:main-comparison}
\vspace{0.5em}
\footnotesize
\setlength{\tabcolsep}{2pt}
\renewcommand{\arraystretch}{1.12}
\resizebox{\textwidth}{!}{%
\begin{tabular}{@{}>{\raggedright\arraybackslash}p{0.165\textwidth}>{\raggedright\arraybackslash}p{0.245\textwidth}ccccc@{}}
\toprule
\textbf{Method} & \textbf{Query-time Model(s)} & \textbf{LVBench} & \multicolumn{2}{c}{\textbf{LongVideoBench}} & \textbf{Video-MME(w/o)} & \textbf{EgoSchema} \\
\cmidrule(lr){3-3}\cmidrule(lr){4-5}\cmidrule(lr){6-6}\cmidrule(lr){7-7}
& & \textbf{Overall} & \textbf{Val} & \textbf{Long} & \shortstack{\textbf{Long}} & \textbf{Val} \\
\midrule
\tableband{groupbeige}
\multicolumn{7}{@{}l}{\makebox[0.41\textwidth][l]{\strut\hspace{0.45em}\textit{Direct Proprietary VLMs}}} \\
Gemini-2.5-Pro & Gemini-2.5-Pro & 72.0 & 71.0 & 68.6 & 75.9 & 72.8 \\
Gemini-3.1-Pro & Gemini-3.1-Pro & 78.2 & 78.6 & 77.0 & 80.3 & 76.4 \\
OpenAI o3 & OpenAI o3 & 57.1 & 66.7 & 60.6 & 64.7 & 63.2 \\
\midrule
\tableband{grouplilac}
\multicolumn{7}{@{}l}{\makebox[0.41\textwidth][l]{\strut\hspace{0.45em}\textit{Direct Open-Source VLMs}}} \\
Seed1.5VL & Seed1.5VL-Thinking & 64.6 & -- & 74.4 & -- & -- \\
Qwen3-VL & Qwen3-VL-235B-A22B-Thinking & 63.6 & 71.4 & 67.1 & 71.2 & 75.9 \\
Qwen3.5 & Qwen3.5-35B-A3B & 71.3 & -- & -- & 62.6 & -- \\
GLM-4.6V & GLM-4.6V-106B-A12B & 59.5 & 67.6 & 58.7 & 66.0 & 68.8 \\
\midrule
\tableband{groupgreen}
\multicolumn{7}{@{}l}{\makebox[0.41\textwidth][l]{\strut\hspace{0.45em}\textit{Agentic Methods}}} \\
DVD & OpenAI o3 & 74.2 & 71.6 & 68.6 & 67.3 & 76.6 \\
Symphony & DeepSeek-R1 + Seed1.6VL & 71.8 & 77.1 & -- & 78.1 & -- \\
VideoARM & o3 + GPT-4.1 & 79.7 & 73.7 & 69.2 & 75.3 & 78.2 \\
VideoARM & o3 + GPT-4o & -- & 78.0 & 76.4 & 81.2 & 76.2 \\
Active Video Perception & Gemini-2.5-Pro & 74.8 & 73.4 & 70.0 & 81.9 & -- \\
WorldMM-GPT & OpenAI GPT-5 & 61.9 & -- & -- & 76.6 & -- \\
VideoSeek & OpenAI GPT-5 & 68.4 & -- & 73.5 & 70.1 & -- \\
LensWalk & OpenAI GPT-5 & 66.9 & -- & 68.8 & 69.2 & 74.6 \\
Route2Look & OpenAI GPT-5 & 75.4 & -- & 77.8 & 76.1 & -- \\
VideoXAgent & Claude-Opus-4.6 + Gemini-3.1-Pro-preview & 76.5 & -- & 78.1 & -- & -- \\
MemDreamer$^{\ddagger}$ & OpenAI GPT-5.5 & 73.3 & 74.1 & 73.9 & 77.1 & 78.6 \\
\midrule
\tableband{groupblue}
\multicolumn{7}{@{}l}{\makebox[0.41\textwidth][l]{\strut\hspace{0.45em}\textit{Same-Backbone End-to-End Comparison}}} \\
GPT-5.5 (Direct)\(^{\dagger}\) & OpenAI GPT-5.5 & 64.3 & 69.8 & 64.0 & 74.8 & 76.0 \\
\tableband{oursblue}
\makebox[0.165\textwidth][l]{\strut\hspace{0.45em}\textbf{VideoTapestry}\(^{\dagger}\)} & \textbf{OpenAI GPT-5.5} & \scoregain{81.5}{17.2} & \scoregain{80.4}{10.6} & \scoregain{78.9}{14.9} & \scoregain{84.6}{9.8} & \scoregain{83.0}{7.0} \\
\bottomrule
\end{tabular}
}
\end{minipage}

\section{Experiments}

\subsection{Benchmarks}

We evaluate VideoTapestry on four long-video understanding benchmarks. LVBench comprises 103 videos ranging from 30 minutes to two hours and 1,549 QA pairs across six fine-grained dimensions~\citep{wang2025lvbench}. The LongVideoBench validation set contains 1,337 questions from 752 videos, including 564 questions from 188 long videos ranging from 15 to 60 minutes~\citep{wu2024longvideobench}. Video-MME (Long) consists of 300 videos ranging from 30 to 60 minutes and 900 QA pairs~\citep{fu2025video}. EgoSchema contains egocentric clips designed to evaluate video reasoning; for efficient and representative evaluation, we use a 500-question subset~\citep{mangalam2023egoschema}.

\subsection{Experimental Setup}

VideoTapestry is training-free, with all model parameters and the preconstructed hierarchical memory fixed during evaluation. Unless noted otherwise, GPT-5.5~\citep{openai2026gpt55} serves as the backbone for the Overview, Skim, Focus, and Answer agents, while Whisper~\citep{radford2023robust} provides ASR. Qwen3-Embedding~\citep{zhang2025qwen3} supports semantic retrieval. The Direct baseline uses the same frozen backbone and evaluation examples without the agentic inference framework, serving as a same-backbone end-to-end comparison. Following VideoARM~\citep{yin2026videoarm}, all ablations are conducted on a fixed 200-question subset of Video-MME Long. Additional evaluation and implementation settings are provided in Appendix~\ref{app:evaluation-settings}.

\paragraph{Baselines.}
We compare with direct VLMs and agentic methods, including DVD~\citep{zhang2026deep}, Symphony~\citep{yan2026symphony}, VideoARM~\citep{yin2026videoarm}, Active Video Perception~\citep{wang2026active}, WorldMM~\citep{yeo2026worldmm}, VideoSeek~\citep{lin2026videoseek}, LensWalk~\citep{li2026lenswalk}, Route2Look~\citep{wang2026routing}, VideoXAgent~\citep{yang2026online}, and MemDreamer~\citep{chen2026memdreamer}.

\subsection{Main Results}

With frozen GPT-5.5, VideoTapestry gains 7.0--17.2 percentage points over the same-backbone Direct baseline and surpasses the strongest prior result in each column by 0.8--4.4 points (Table~\ref{tab:main-comparison}). It also leads across all six LVBench dimensions, with the largest same-backbone gains in temporal grounding (+40.1) and summarization (+24.3) (Table~\ref{tab:lvbench-breakdown}). These results suggest benefits for both long-range evidence localization and global information aggregation.

\begin{table*}[t]
\centering
\caption{Comparison with representative VLMs and agentic approaches across the six capability dimensions of LVBench (accuracy, \%). The best results are bolded; green parenthesized values denote absolute percentage-point gains over GPT-5.5 (Direct). \(\ddagger\) denotes results from our reproduction.}
\label{tab:lvbench-breakdown}
\vspace{0.5em}
\small
\settowidth{\lvbenchpendingwidth}{\textit{Pending}}
\setlength{\tabcolsep}{1.0pt}
\renewcommand{\arraystretch}{1.05}
\resizebox{\textwidth}{!}{%
\begin{tabular}{lccccccc}
\toprule
\textbf{Method}
& \textbf{ER}
& \textbf{EU}
& \textbf{Rea}
& \textbf{KIR}
& \textbf{Sum}
& \textbf{TG}
& \textbf{Avg.} \\
\midrule
\tableband{groupbeige}
\multicolumn{8}{l}{\makebox[0.28\textwidth][l]{\strut\hspace{0.45em}\textit{Direct VLMs}}} \\
GPT-4o
& 48.9 & 49.5 & 50.3 & 48.1 & 50.0 & 40.9 & 48.9 \\
OpenAI o3
& 57.6 & 56.4 & 50.8 & 62.9 & 67.2 & 46.8 & 57.1 \\
GPT-5.5 (Direct)
& \makebox[\lvbenchpendingwidth][c]{63.2}
& \makebox[\lvbenchpendingwidth][c]{66.0}
& \makebox[\lvbenchpendingwidth][c]{72.7}
& \makebox[\lvbenchpendingwidth][c]{63.8}
& \makebox[\lvbenchpendingwidth][c]{65.4}
& \makebox[\lvbenchpendingwidth][c]{43.1}
& \makebox[\lvbenchpendingwidth][c]{64.3} \\
\midrule
\tableband{groupgreen}
\multicolumn{8}{l}{\makebox[0.28\textwidth][l]{\strut\hspace{0.45em}\textit{Structured and Agentic Methods}}} \\
VideoTree
& 30.3 & 25.1 & 31.9 & 26.5 & 25.5 & 27.7 & 28.8 \\
VideoLucy
& 54.3 & 59.8 & 55.9 & 75.6 & 49.1 & 51.7 & 58.8 \\
DVD
& 73.4 & 73.3 & 70.7 & 80.4 & 74.1 & 72.3 & 74.2 \\
DVD + Transcripts
& 75.5 & 77.1 & 68.7 & 79.0 & 84.5 & 72.7 & 76.0 \\
Symphony
& 70.0 & 69.4 & 69.4 & 77.2 & 72.5 & 70.1 & 71.8 \\
Active Video Perception
& 71.9 & 76.7 & 67.7 & 80.1 & 75.9 & 73.6 & 74.8 \\
MemDreamer (GPT-5.5)$^{\ddagger}$ 
& 70.5 & 74.9 & 73.5 & 80.1 & 67.3 & 67.7 & 73.3 \\
\midrule
\tableband{oursblue}
\makebox[0.28\textwidth][l]{\strut\hspace{0.45em}\textbf{VideoTapestry (GPT-5.5)}}
& \scoregain{80.7}{17.5} & \scoregain{81.0}{15.0} & \scoregain{77.6}{4.9}
& \scoregain{84.5}{20.7} & \scoregain{89.7}{24.3} & \scoregain{83.2}{40.1}
& \scoregain{81.5}{17.2} \\
\bottomrule
\end{tabular}%
}
\end{table*}

The LVBench breakdown in Table~\ref{tab:lvbench-breakdown} shows that VideoTapestry leads all six capability dimensions, improving over the strongest prior result by 3.9--9.6 points. The largest gains occur in temporal grounding (+9.6), supporting the effectiveness of coarse-to-fine retrieval and targeted observation for locating and verifying question-relevant evidence.

\subsection{Ablation Studies}

Following VideoARM~\citep{yin2026videoarm}, all controlled studies use its representative 200-question subset of Video-MME Long, which was constructed to preserve the distribution of the full benchmark, and share the same preconstructed memory.

\paragraph{Query-time memory refinement.}
Table~\ref{tab:ablations}(a) examines query-specific evidence acquisition and integration into hierarchical memory. The two observation ablations remove visual observation or selected-window ASR. \textit{Append-only Evidence} retains the same tools and agent structure but provides new evidence as a flat log without updating selected memory nodes. \textit{Single Agent} replaces the three scale-aligned agents with one controller under the same model- and tool-call budgets. Removing visual observation, ASR, structured memory refinement, and multi-agent collaboration reduces accuracy by 3.0, 5.5, 1.5, and 4.5 points, respectively. The largest drops come from removing ASR and scale-aligned collaboration, while structured memory refinement outperforms flat evidence logging.

\paragraph{Composite memory.}
Table~\ref{tab:ablations}(b) separates the static-retrieval baseline from a controlled comparison of composite memory.
\textit{Static Memory Only} retrieves the preconstructed memory without query-time updates, compression, or rewriting, achieving 79.5\%.
For the refined variants, the full exploration is fixed, including selected nodes, observations, refined contents, and query-time trace; only the final memory given to the Answer Agent differs: \textit{Refined Super Only} keeps the refined Super-level scaffold (85.0\%), while \textit{Refined Paths Only} keeps only refined Macro- and Subgraph-level paths (80.5\%).
Combining both components reaches 87.0\%, outperforming either refined component and static retrieval by 7.5 percentage points, supporting their complementary roles.

\begin{table*}[t]
\centering
\caption{Ablations of (a) query-time refinement and (b) composite memory components. Update denotes structured integration; M-A, scale-aligned multi-agent collaboration. $\Delta$ Acc. denotes the percentage-point difference from VideoTapestry. In (b), refined variants share fixed exploration; Super/Paths denote retained components, and -- static retrieval.}
\label{tab:ablations}
\vspace{0.5em}
\small

\textbf{(a) Query-time memory refinement}\\[2pt]
{\setlength{\tabcolsep}{3.5pt}%
\begin{tabular*}{0.84\textwidth}
{@{\extracolsep{\fill}}lcccccc@{}}
\toprule
Variant & Visual & ASR & Update & M-A & Acc. & $\Delta$ Acc. \\
\midrule
w/o Visual Observation
& N & Y & Y & Y
& 84.0 & $-3.0$ \\

w/o ASR Observation
& Y & N & Y & Y
& 81.5 & $-5.5$ \\

Append-only Evidence
& Y & Y & N & Y
& 85.5 & $-1.5$ \\

Single Agent
& Y & Y & Y & N
& 82.5 & $-4.5$ \\

\tableband{oursblue}
\textbf{VideoTapestry}
& Y & Y & Y & Y
& \textbf{87.0} & -- \\
\bottomrule
\end{tabular*}}

\vspace{4pt}

\textbf{(b) Composite Memory Components}\\[2pt]
{\setlength{\tabcolsep}{6pt}%
\begin{tabular*}{0.84\textwidth}
{@{\extracolsep{\fill}}lcccc@{}}
\toprule
Memory Variant & Super & Paths & Tokens & Acc. \\
\midrule
Static Memory Only
& -- & --
& 4,874 & 79.5 \\

Refined Super Only
& Y & N
& 3,250 & 85.0 \\

Refined Paths Only
& N & Y
& 1,363 & 80.5 \\

\tableband{oursblue}
\textbf{VideoTapestry}
& Y & Y
& 4,417 & \textbf{87.0} \\
\bottomrule
\end{tabular*}}
\vspace{-2mm}
\end{table*}

\subsection{Backbone Generalization}

\noindent
\begin{minipage}[t]{0.48\textwidth}
\vspace{0pt}
Table~\ref{tab:backbone-generalization} evaluates backbone generalization while keeping the memory and tool interfaces fixed. We replace the backbone used by the agents and Answer Agent with GPT-5.5, GPT-5, Qwen3.7-Plus, and Qwen3.6-27B~\citep{openai2025gpt5,alibabacloud2026qwen37plus,qwen2026qwen36}. Direct and VideoTapestry use the same backbone and evaluation examples. VideoTapestry improves every backbone by 7.0--9.5\%, with the largest gain of 9.5\% on Qwen3.6-27B. These consistent gains across GPT and Qwen show the method generalizes beyond a single model family.
\end{minipage}
\hfill
\begin{minipage}[t]{0.49\textwidth}
\vspace{0pt}
\centering
\captionof{table}{Generalization across backbone models.}
\label{tab:backbone-generalization}
\small
\setlength{\tabcolsep}{2pt}
\vspace{5pt}
\begin{tabular*}{\linewidth}{@{\extracolsep{\fill}}lccc@{}}
\toprule
Backbone & Direct & VideoTapestry & Improve. \\
\midrule
GPT-5.5
& 78.0 & 87.0 & $+9.0$ \\
GPT-5
& 75.5 & 82.5 & $+7.0$ \\
Qwen3.7-Plus
& 75.0 & 83.0 & $+8.0$ \\
Qwen3.6-27B
& 65.5 & 75.0 & $+9.5$ \\
\bottomrule
\end{tabular*}
\end{minipage}

\section{Conclusion}

We introduce VideoTapestry, a training-free hierarchical multi-agent framework for long-video understanding. Starting from a query-independent hierarchical video memory, the Overview, Skim, and Focus agents perform coarse-to-fine retrieval and targeted observation, refining query-relevant content at complementary semantic granularities. Their layer-wise refinements are assembled with the global scaffold into a composite query-adaptive memory that preserves video-wide structure while enriching relevant branches for final reasoning. Experiments on long-video benchmarks show that VideoTapestry consistently improves over the same-backbone direct baseline and representative agentic methods, demonstrating the effectiveness of organizing global structure and query-specific evidence for final reasoning.

\bibliography{references}
\bibliographystyle{iclr2027_conference}

\appendix

\clearpage

\section{Additional Results}

\paragraph{Additional Qualitative Results.}
Figure~\ref{fig:qualitative-cases} shows two representative trajectories from coarse localization to fine-grained verification, illustrating progressive query-adaptive memory refinement.

\begin{center}
\centering
\begin{minipage}{0.90\textwidth}
\raggedright
\textbf{Case 1: Fine-grained temporal verification.}
\par\smallskip
The agents narrow the stairwell interval to 27--33 s, verify descent without door or
window opening, and answer C.
\end{minipage}

\vspace{0.25em}

\includegraphics[width=0.90\textwidth]{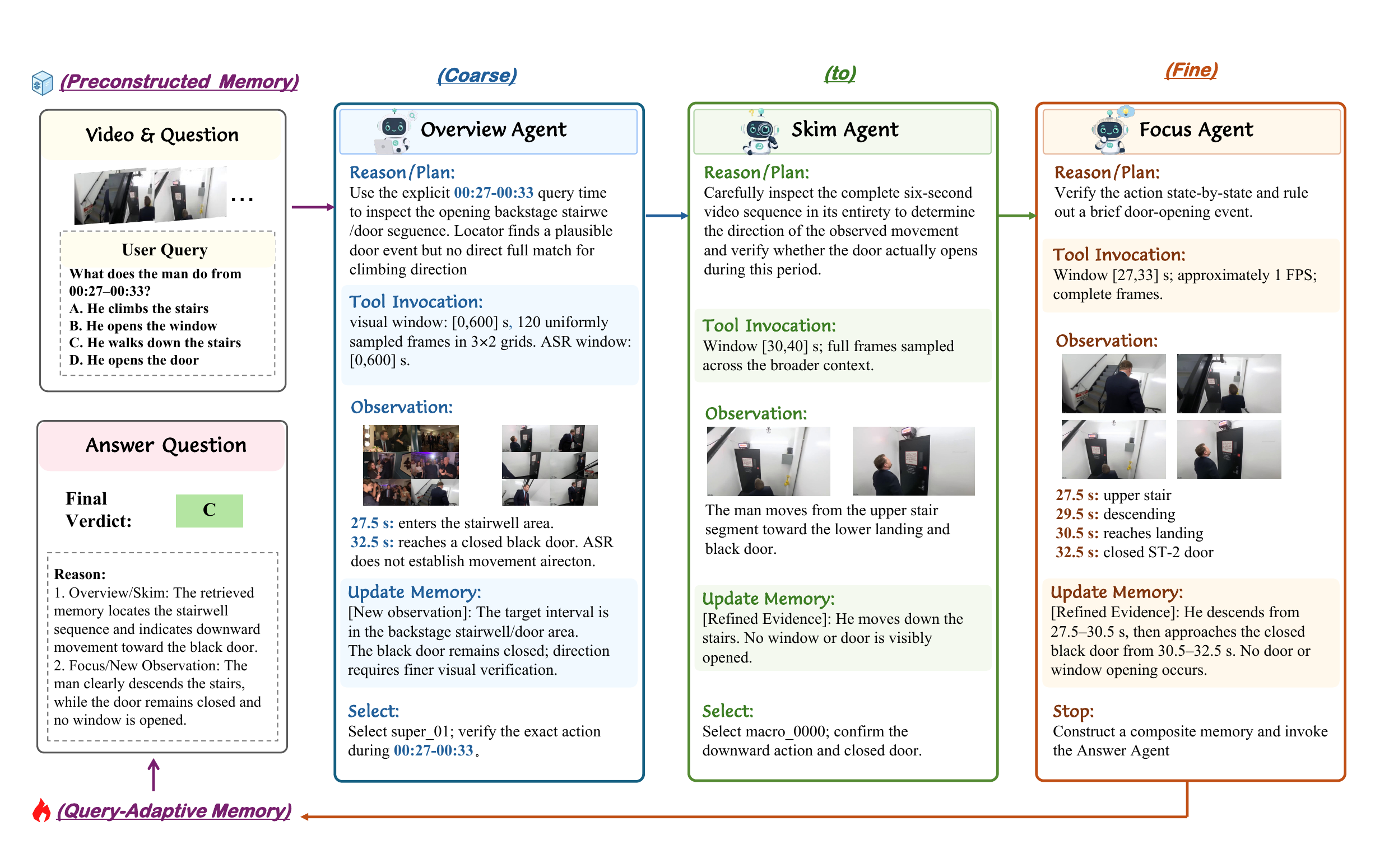}

\vspace{0.5em}

\begin{minipage}{0.90\textwidth}
\raggedright
\textbf{Case 2: Cross-branch evidence aggregation.}
\par\smallskip
The agents verify distinct performers across two piano-related branches roughly
25 minutes apart, aggregate the evidence, and answer D.
\end{minipage}

\vspace{0.25em}

\includegraphics[width=0.90\textwidth]{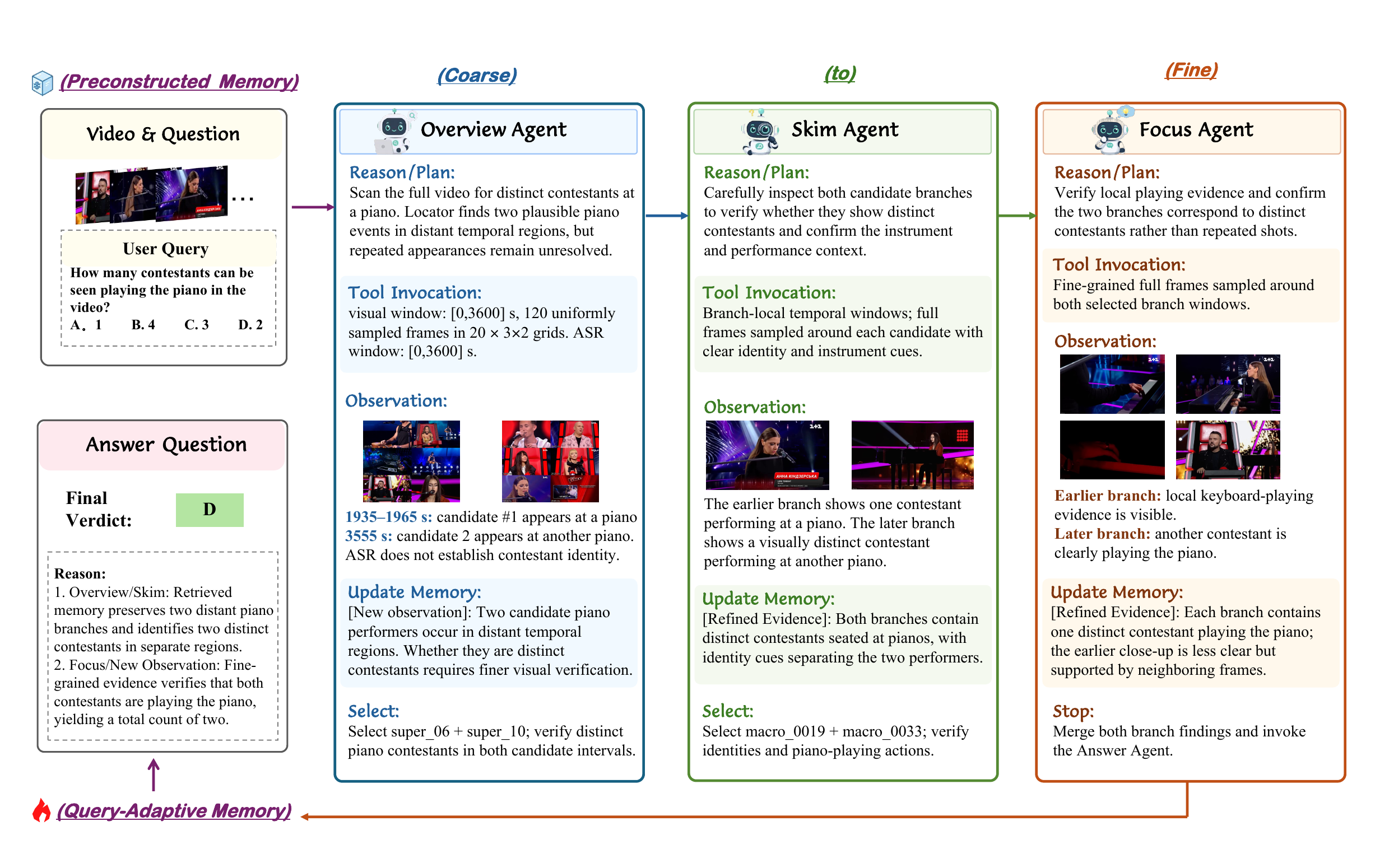}

\captionof{figure}{Two examples of coarse-to-fine query-adaptive memory refinement.}
\label{fig:qualitative-cases}
\end{center}

\clearpage

\section{Experimental and Inference Details}

\subsection{Evaluation and Implementation Settings}
\label{app:evaluation-settings}

\paragraph{Closed- and open-source baseline evaluation.}
Values in the main table are obtained from the corresponding models' technical reports or previously published benchmark results when available; the remaining entries are obtained from our evaluations. GPT- and Qwen-based models are evaluated via their corresponding APIs across all four benchmarks. Direct baselines receive uniformly sampled video frames. Under the maximum image-upload limits of the API gateways used in our experiments, GPT-family models receive 50 frames, Qwen3.7-Plus receives 196 frames, and Qwen3.6-27B receives 396 frames per video. For GPT-family models, the reasoning effort is set to high to enable extended reasoning, while Qwen-family models use thinking mode during direct inference. All inference runs use a temperature of 0.1.

\paragraph{VideoTapestry inference setup and memory construction.}
Each Visual Observation Tool call uploads at most 50 images, respecting the GPT API gateway limit. Overview samples up to 120 frames and packs them into at most 20 six-frame grids, whereas Skim and Focus each use up to 50 individual frames from selected intervals. For fair comparison, the ASR Tool applies Whisper only to agent-selected audio windows rather than transcribing the full video. For GPT-family models, only the Answer Agent uses high reasoning effort, while all other calls use the default setting. For Qwen-based experiments, thinking is enabled only for planning, decision, and final answering, while all other calls use non-thinking mode. To isolate our query-time method from offline memory construction, all experiments use hierarchical video memories built following the MemDreamer procedure~\citep{chen2026memdreamer}. For the MemDreamer baseline, we separately reproduce its retrieval pipeline.

\paragraph{ASR Tool implementation details.}
The ASR Tool provides query-dependent access to spoken information within agent-selected temporal windows. For benchmarks without provided subtitles, including Video-MME (w/o subtitles), no subtitle text or full-video transcript is supplied to the model. After the Overview Agent identifies relevant temporal regions from the query and visual evidence, Whisper is applied only to the selected windows. At most three windows are accessed, each normalized to 600--1200\,s within the video bounds, rather than transcribing the entire video. For benchmarks with provided subtitles, the same temporal-selection procedure is retained, but the ASR Tool directly retrieves subtitle text from the selected windows instead of invoking Whisper. In both cases, spoken information is exposed only on demand over agent-selected regions rather than provided globally throughout inference.

\paragraph{Visual observation and grid-packing protocol.}
Direct baselines receive uniformly sampled individual video frames without grid packing. We retain this setting because grid packing is part of VideoTapestry's coarse observation mechanism rather than a generic preprocessing step; applying it to Direct would introduce an additional temporal-compression strategy beyond standard frame prompting. Within VideoTapestry, the Overview Agent samples up to 120 frames and packs them into at most 20 six-frame grids, providing broader temporal coverage while remaining within the per-call image limit described above. Once candidate regions are localized, the Skim and Focus Agents instead use individual frames from selected intervals to preserve finer visual detail for subsequent observation.

\clearpage

\subsection{Detailed Inference Pipeline}
\label{app:inference_details}

Algorithms~1 and~2 jointly describe VideoTapestry's inter-level coordination and intra-level refinement. Algorithm~1 summarizes the end-to-end hierarchical inference, including candidate propagation, instruction handoff, stopping, and composite memory assembly. Algorithm~2 specifies the per-layer Plan--Observe--Update--Decide procedure used across all four benchmarks, with one observation/refinement batch and one decision per visited layer. Overview performs coarse visual observation with selected-window ASR, while Skim and Focus conduct progressively finer visual observation.

\vspace{0.5em}
\noindent
\begin{minipage}{\textwidth}
\footnotesize
\hrule
\vspace{0.4em}
\textbf{Algorithm 1: Inference Pipeline of VideoTapestry}
\vspace{0.35em}
\hrule
\vspace{0.4em}
\setlength{\tabcolsep}{3pt}
\renewcommand{\arraystretch}{1.08}

\begin{tabular}{@{}>{\bfseries}r p{0.88\textwidth}@{}}
Input: & Video $V$, question $q$, options $o$, static hierarchy
$\mathcal{G}$ and memory $\mathcal{M}_{\mathrm{static}}$,
refinement agents $\{\mathcal{A}_{S},\mathcal{A}_{M},
\mathcal{A}_{G}\}$, and Answer Agent $\mathcal{A}_{\mathrm{ans}}$. \\
Output: & Predicted answer $\widehat{y}$. \\
\end{tabular}
\vspace{0.2em}

\begin{tabular}{@{}r@{\hspace{0.6em}}p{0.92\textwidth}@{}}
1: & Initialize $\mathcal{C}^{S}$ with all SuperEvents,
$\mathcal{I}^{0}\leftarrow\varnothing$,
$\mathcal{L}_{q}\leftarrow\varnothing$, and trace
$H_q\leftarrow[\,]$. \\
2: & \textbf{for} $\ell\in\{S,M,G\}$, aligned with
Overview, Skim, and Focus, \textbf{do} \\
3: & \hspace{1em}Load original candidate memory
$\mathcal{M}^{\ell}\leftarrow
\mathcal{M}_{\mathrm{static}}^{\ell}[\mathcal{C}^{\ell}]$. \\
4: & \hspace{1em}\textbf{if} $\mathcal{M}^{\ell}=\varnothing$,
\textbf{break}. \\
5: & \hspace{1em}Run Algorithm~2 to obtain
$(\widetilde{\mathcal{M}}_{q}^{\ell},
\mathcal{S}^{\ell},\mathcal{I}^{\ell},z^{\ell})$. \\
6: & \hspace{1em}$\mathcal{L}_{q}\leftarrow
\mathcal{L}_{q}\cup\{\ell\}$; append the agent's handoff
instruction or stopping marker to $H_q$. \\
7: & \hspace{1em}\textbf{if} $z^{\ell}$ is true,
\textbf{break}. \\
8: & \hspace{1em}$\mathcal{C}^{\ell+1}\leftarrow
\bigcup_{v\in\mathcal{S}^{\ell}}
\operatorname{Children}(v,\mathcal{G})$. \\
9: & \textbf{end for} \\
10: & $\mathcal{D}_{q}\leftarrow
\operatorname{Serialize}\!\left(
q,o,H_q,
\operatorname{Assemble}\!\left(
\{\widetilde{\mathcal{M}}_{q}^{\ell}\}_{\ell\in\mathcal{L}_{q}}
\right)\right)$. \\
11: & Return $\widehat{y}\leftarrow
\mathcal{A}_{\mathrm{ans}}(q,o,\mathcal{D}_{q})$
without further retrieval or observation. \\
\end{tabular}
\vspace{0.35em}
\hrule
\end{minipage}

\vspace{0.8em}
\noindent
\begin{minipage}{\textwidth}
\footnotesize
\hrule
\vspace{0.4em}
\textbf{Algorithm 2: Layer-Wise Observation and Memory Refinement}
\vspace{0.35em}
\hrule
\vspace{0.4em}
\setlength{\tabcolsep}{3pt}
\renewcommand{\arraystretch}{1.08}

\begin{tabular}{@{}>{\bfseries}r p{0.88\textwidth}@{}}
Input: & Level $\ell$, video $V$, question $q$, options $o$,
nonempty static candidate memory $\mathcal{M}^{\ell}$,
and upstream instruction $\mathcal{I}^{\ell-1}$. \\
Output: & Refined memory $\widetilde{\mathcal{M}}_{q}^{\ell}$,
selected IDs $\mathcal{S}^{\ell}$, downstream instruction
$\mathcal{I}^{\ell}$, and stop flag $z^{\ell}$. \\
\end{tabular}
\vspace{0.2em}

\begin{tabular}{@{}r@{\hspace{0.6em}}p{0.92\textwidth}@{}}
1: & Initialize local trajectory $H_{\ell}\leftarrow[\,]$
and observations $O\leftarrow\varnothing$;
$g\leftarrow\operatorname{GlobalHeuristic}(q,o)$. \\
2: & $p\leftarrow\operatorname{Plan}_{\ell}
(q,o,\mathcal{M}^{\ell},\mathcal{I}^{\ell-1},H_{\ell})$;
at $S$, include locator hints. \\
3: & \textbf{if} $\ell=S$ \textbf{then} \\
4: & \hspace{1em}Merge planned ranges into one video-bounded
window; fall back to the full video if invalid, and expand
to at least $\min(600\,\mathrm{s},\operatorname{duration}(V))$. \\
5: & \hspace{1em}$O_v\leftarrow\operatorname{VisualObserve}$
using up to 120 frames packed into 20 six-frame grids. \\
6: & \hspace{1em}$p_a\leftarrow\operatorname{PlanASR}
(q,o,\mathcal{M}^{S},p,O_v)$; fall back to the visual
window if no ASR ranges are proposed. \\
7: & \hspace{1em}Normalize up to three ASR windows to
600--1200\,s each within video bounds
(use the full video if shorter than 600\,s). \\
8: & \hspace{1em}$O_a\leftarrow\operatorname{ASRObserve}(p_a)$
using Whisper transcripts;
$O\leftarrow\{O_v,O_a\}$. \\
9: & \hspace{1em}Construct $\mathcal{P}$ from evidence-linked
SuperEvents and their valid immediate temporal neighbors. \\
10: & \hspace{1em}Build $\widetilde{\mathcal{M}}_{q}^{S}$:
retain original text in $\mathcal{P}$ and append available
associated evidence; compress nodes outside $\mathcal{P}$. \\
11: & \textbf{else} \\
12: & \hspace{1em}\textbf{if}
$p.\mathrm{observation\_needed}$ or
$(\ell=M\ \mathrm{and}\ \neg g)$ \textbf{then} \\
13: & \hspace{2em}Intersect planned ranges with candidate
intervals (fallback: candidate intervals);
$O\leftarrow\operatorname{VisualObserve}$
using up to 50 individual frames. \\
14: & \hspace{1em}\textbf{end if};
$\widetilde{\mathcal{M}}_{q}^{\ell}\leftarrow
\operatorname{Update}_{\ell}
(q,o,\mathcal{I}^{\ell-1},
\mathcal{M}^{\ell},p,O)$ for valid requested nodes. \\
15: & \textbf{end if}; record plans, observations, and
updates in $H_{\ell}$. \\
16: & $(\mathcal{S}^{\ell},\mathcal{I}^{\ell},z^{\ell})
\leftarrow\operatorname{Decide}
(q,o,\mathcal{M}^{\ell},
\widetilde{\mathcal{M}}_{q}^{\ell},
\mathcal{I}^{\ell-1},H_{\ell})$ once. \\
17: & Force $z^{\ell}\leftarrow\mathrm{true}$ at $G$;
at $S$ or $M$, force $z^{\ell}\leftarrow\mathrm{false}$
for non-global questions. Validate selections and apply
fallback IDs/instructions when continuing; stop if no
valid descendants remain. \\
18: & If stopping, clear $\mathcal{S}^{\ell}$ and
$\mathcal{I}^{\ell}$; return
$(\widetilde{\mathcal{M}}_{q}^{\ell},
\mathcal{S}^{\ell},\mathcal{I}^{\ell},z^{\ell})$. \\
\end{tabular}
\vspace{0.35em}
\hrule
\end{minipage}

\clearpage

\subsection{Prompts Used in VideoTapestry}

We provide the core prompts used throughout our framework, covering the main stages of the overall pipeline and the key components involved in its execution, including planning, observation, memory updating, decision making, and answer generation.

\subsubsection{Hierarchical Observation Planning}

\begin{agentpromptbox}{Overview Agent Planning Prompt}{overviewblue}
YOU ARE OVERVIEW AGENT.
You are responsible for updating the current-level hierarchical video memory under a specific question.
You can inspect visual frames through OBSERVATION TOOL and rewrite old memory text through UPDATE TOOL.
Your goal is to produce concise, accurate, query-relevant dynamic memory for the current layer.
Old memory text is only a static prebuilt summary. It is not visual verification and must not be treated as a substitute for observation. Only set observation_needed=false when the video file or visual frames are unavailable, or when this layer already has a prior visual observation in trajectory.
For Overview/Super, observation_needed must be true for the first update attempt.
Decide one bounded observation/update batch. Do not answer the question.
Return strict JSON with keys: thought, observation_needed, observation_instruction, candidate_memory_ids, time_ranges, locator_time_hints_used.
For Overview/Super, return exactly one contiguous visual observation time range inside observation_time_bounds. Center it on the evidence region you need to inspect; it may extend beyond an individual Super boundary. Choose a span of at least 600 seconds when the video duration permits; the program enforces the final minimum span.
overview_locator_results, when present, are only global time-localization leads from entity/OCR/node search. Do not treat their text as visual facts or as support for an option. Use their timestamps only to choose a question-relevant continuous observation window, then require Observation Tool to verify the fact visually.
When tool=multi_locator_compressed, candidate_windows are complete joint hypotheses containing one grounded match per atomic anchor. Prefer a complete, exact, temporally compact cluster; do not substitute an earlier partial look-alike merely because it ranks highly. Expand the chosen joint time_range into the required continuous observation span. If candidate_windows is empty or unresolved_anchors is nonempty, do not narrow the observation from partial Locator hits; instead use the complete Super memories to choose a broader window. Preserve plausible competing complete windows in locator_time_hints_used. The compressed result remains unverified localization, not answer evidence.
Return locator_time_hints_used as a list of the temporal hints you actually adopted. Each item must explain only why a time window was selected; do not copy raw search output, query text, option language, or unverified factual claims. Return [] if none was used.
candidate_memory_ids must come from supplied memory_id values.
Current level is super. Respect the tool limits: at most 2 observations, at most 2 updates, at most 4 total calls.
\end{agentpromptbox}

\begin{agentpromptbox}{Skim / Focus Planning Template}{sharedpurple}
YOU ARE {AGENT_NAME}.
You are responsible for updating the current-level hierarchical video memory under a specific question.
You can inspect visual frames through OBSERVATION TOOL and rewrite old memory text through UPDATE TOOL.
Your goal is to produce concise, accurate, query-relevant dynamic memory for the current layer.
Old memory text is only a static prebuilt summary. It is not visual verification and must not be treated as a substitute for observation. Only set observation_needed=false when the video file or visual frames are unavailable, or when this layer already has a prior visual observation in trajectory.
For Overview/Super, observation_needed must be true for the first update attempt.
Decide one bounded observation/update batch. Do not answer the question.
Return strict JSON with keys: thought, observation_needed, observation_instruction, candidate_memory_ids, time_ranges, update_memory_ids.
time_ranges must stay inside the time_range of the supplied old_memories.
candidate_memory_ids and update_memory_ids must come from supplied memory_id values.
Current level is {MEMORY_LEVEL}. Respect the tool limits: at most 2 observations, at most 2 updates, at most 4 total calls.
\end{agentpromptbox}

\newpage

\subsubsection{Visual Observation Tool}

\begin{agentpromptbox}{Visual Observation Tool Prompt}{sharedpurple}
YOU ARE OBSERVATION TOOL.
You are being called by {AGENT_NAME} for the {MEMORY_LEVEL} layer.
Inspect the provided visual frames and answer only the supplied observation_instruction. Do not update memory, infer an unstated question, or choose a multiple-choice answer.
For observation_result: describe specific visible facts (people, actions, objects, on-screen text, timestamps shown). Include only facts that directly address observation_instruction.
For uncertainty: list only facts you are unsure about.
For relevance_to_question: one sentence summary of how the frames answer observation_instruction.
If a fact is unclear, write it as uncertain instead of a confirmed fact.
Return strict JSON with keys: observation_result, uncertainty, relevance_to_question.
\end{agentpromptbox}

\begin{agentpromptbox}{ASR Observation Tool Prompt}{overviewblue}
YOU ARE THE OVERVIEW AGENT'S ASR OBSERVATION TOOL.

You will receive exactly two inputs:
1. observation_instruction: the self-contained spoken-information request;
2. asr: the timestamped ASR transcript from the selected video window

You do not receive the original question or answer options. Analyze only the
supplied ASR to answer observation_instruction with objective facts.

OUTPUT REQUIREMENTS

1. observation_result
- Answer observation_instruction using only facts supported by the ASR.
- When evidence is conclusive, state the underlying fact clearly.
- When the evidence is incomplete, conflicting, or unclear, explicitly describe
  what cannot be determined.
- Do not use outside knowledge and do not select a multiple-choice answer.
2. asr_evidence
- Copy the most important supporting ASR lines exactly.
- Preserve timestamps, text, punctuation, capitalization, line breaks, and
  transcription errors.
- Do not paraphrase, translate, summarize, correct, merge, reorder, or rewrite.
- Include only directly supporting lines, in chronological order.
- Every character must come from the supplied asr.
- Return an empty string when there is no reliable evidence.

Return strict JSON with exactly these two fields:

{
  "observation_result": "...",
  "asr_evidence": "..."
}
\end{agentpromptbox}

\newpage

\subsubsection{Memory Update Tool}

\begin{agentpromptbox}{SuperEvent Update Prompt}{overviewblue}
YOU ARE UPDATE TOOL.

You are being called by Overview Agent for the super layer.

You receive question, options, previous_instruct, one shared update_instruction,
observation_sources, and update_requests. Each long observation source appears
exactly once in observation_sources. Every update request contains memory_id,
old_memory_text, and source_refs. Resolve source_refs against observation_sources.
Different Super memories may reference the same source so they can extract
different, time-local or narrative-context facts without duplicating the source
text in the request. Do not merge visual and ASR observations.

Do not modify memory_id, time_range, level, or parent_id.

SUPER LAYER RULES:

- First set is_relevant to true or false.

- If is_relevant is true:
  1. Write only concise objective visual facts from referenced visual sources
     into visual_observation. If no visual source is referenced, return empty.
  2. Write only concise objective spoken facts from referenced ASR sources into
     asr_observation. If no ASR source is referenced, return empty.
  3. Keep the two modalities separate. Do not move or duplicate facts between them.
  4. Do not repeat, summarize, rewrite, or quote old_memory_text.
  5. Do not return or quote raw asr_evidence. The program assigns original ASR
     lines to memories from their timestamps after this call.
  6. Keep irrelevant_summary empty.

- If is_relevant is false:
  1. Keep visual_observation and asr_observation empty.
  2. Write a 1-2 sentence objective summary of the segment's topic and
     key content in irrelevant_summary.
  3. Do not explain why the segment is irrelevant.

Do not include a "[Visual observation]", "[ASR observation]", or
"[ASR evidence]" heading. The program adds these headings when writing memory.

STRICT PROHIBITION:

Never write any option-tendency judgment in memory text.
Never write phrases such as "consistent with B", "supports option A",
"more likely X", "the correct answer is", or any equivalent statement
that steers toward a specific answer choice.

Only state objective facts supported by old memory or the supplied
observation. The final answer agent will judge the options.

Do not invent information that is not supported by old_memory_text or referenced
observation sources.

Preserve every supplied memory_id exactly.

Return strict JSON exactly in this schema:

{
  "updated_memories": {
    "memory_id": {
      "is_relevant": true,
      "visual_observation": "...",
      "asr_observation": "...",
      "irrelevant_summary": ""
    }
  }
}
\end{agentpromptbox}

\begin{agentpromptbox}{MacroEvent / Subgraph Update Prompt}{sharedpurple}
YOU ARE UPDATE TOOL.
You are being called by {AGENT_NAME} for the {MEMORY_LEVEL} layer.
Read the old memory text, question, options, and trajectory observations. Do not modify memory_id, time_range, level, or parent_id.
{MEMORY_LEVEL_UPPERCASE} LAYER RULES:
- RELEVANT memory: preserve specific facts - visible people, actions, objects, on-screen text, timestamps. Do not replace concrete details with vague phrases.
- IRRELEVANT memory: write an extremely short summary (<=10 characters).
STRICT PROHIBITION: Never write any option-tendency judgment in text. Never write phrases like 'consistent with B', 'supports option A', 'more likely X', or any statement that steers toward a specific answer choice. Only state observed objective facts. The final answer agent will judge the options.
Do not invent information not supported by old memory or observations.
Return strict JSON exactly in this schema: {"updated_memories":{"memory_id":{"updated_text":"..."}}}.
\end{agentpromptbox}

\newpage

\subsubsection{Hierarchical Decisions and Inter-Agent Handoff}

\begin{agentpromptbox}{Shared Decision Prompt}{sharedpurple}
YOU ARE {AGENT_NAME}.
Decide whether the updated current-level dynamic memory is sufficient or whether to descend. Do not answer the question.
{LEVEL_SPECIFIC_INSTRUCTION}
Return strict JSON with keys: selected_child_memory_ids, stop_descending, next_instruct, reason, template_type. If stop_descending is true, selected_child_memory_ids must be [] and next_instruct must be null. If continuing, selected_child_memory_ids must be current-level memory ids supplied in updated_dynamic_memories.
\end{agentpromptbox}

\begin{agentpromptbox}{Hierarchical Decision Instructions}{sharedpurple}
(*@\textcolor{overviewblue}{\ttfamily\bfseries\footnotesize If $\ell=L_0$ (Super), decide $L_0 \rightarrow L_1$:}@*)
If continuing to Macro, generate next_instruct using one of these L0->L1 types: main_idea, temporal_order, identity, action_verification, spatial_relation, causal_result. The next_instruct must be one sentence that tells Skim Agent what type of Macro information to find. Return template_type as the selected type. Only global main-idea/title/summary questions may stop at Super. All other question types must continue to Macro.

(*@\textcolor{skimgreen}{\ttfamily\bfseries\footnotesize Else if $\ell=L_1$ (Macro), decide $L_1 \rightarrow L_2$:}@*)
If continuing to Subgraph, generate a short instruction in the form: Found: [...]; still need to verify: [...]. For non-global detail questions, prefer continuing to Subgraph after at least one Macro observation; old Macro text alone is not enough to terminate.

(*@\textcolor{focusbrown}{\ttfamily\bfseries\footnotesize Else if $\ell=L_2$ (Subgraph), return $\mathrm{STOP}$:}@*)
Focus is the final retrieval layer; stop_descending must be true.
\end{agentpromptbox}

\subsubsection{Final Answer Agent}

\begin{agentpromptbox}{Final Answer Agent System Prompt}{answergray}
YOU ARE THE FINAL ANSWER AGENT.
Based on the memory provided below, return strict JSON with keys: reasoning, answer. 
\end{agentpromptbox}

\begin{agentpromptbox}{Final Answer Agent User Prompt}{answergray}
Question:
{RAW_QUESTION_WITH_OPTIONS}

Retrieved video memory:
{SERIALIZED_QUERY_ADAPTIVE_MEMORY}
\end{agentpromptbox}

\clearpage

\section{Memory Ablation Protocol}
\label{app:memory_ablation}

\paragraph{Shared evaluation setting.}
All configurations in Table~\ref{tab:ablations}(b) use the same
200-question subset of Video-MME Long, GPT-5.5, and the same
preconstructed hierarchical memory. We distinguish the static-retrieval
baseline from the controlled comparison of final memory components.

\paragraph{Static retrieval baseline.}
\textit{Static Memory Only} performs hierarchical retrieval without
Visual, ASR, or Update calls. The Answer Agent receives the retrieved
static SuperEvent, MacroEvent, and Subgraph contents together with the
retrieval trace from that run; node contents are neither compressed nor
rewritten. Its comparison with VideoTapestry therefore measures the
combined effect of query-time observation and memory refinement rather
than memory composition alone.

\paragraph{Controlled final-memory ablations.}
For each question, VideoTapestry first completes its full query-time
exploration. The resulting selected nodes, observations, refined
contents, and exploration trace are then fixed, and three final memory
representations are constructed:
\begin{itemize}
    \item \textit{Refined Super Only} retains the refined Super-level
    scaffold while excluding MacroEvent and Subgraph contents.
    \item \textit{Refined Paths Only} retains refined MacroEvent and
    Subgraph contents along selected paths while excluding SuperEvents.
    \item \textit{VideoTapestry} combines the refined Super-level
    scaffold with the refined paths.
\end{itemize}
The three variants use the same exploration trace and do not rerun
retrieval, observation, or updating; only the memory contents supplied
to the Answer Agent differ. Thus, their comparison isolates the
contribution of the final serialized memory components under a fixed
exploration result.

\section{Limitations and Future Work}
\label{app:limitations}

VideoTapestry is evaluated with a fixed, preconstructed video hierarchy. Although query-time observations can supplement missing details, retrieval remains constrained by the initial segmentation and parent--child structure. Incorrect event grouping or temporal boundaries may therefore hinder access to relevant evidence. Evaluating the framework with alternative memory constructors and allowing selective structural revision are promising directions.

The framework also incurs offline memory-construction costs and additional query-time model and tool calls. The reported accuracy gains therefore do not establish a computational-efficiency advantage. Future work could investigate budget-aware observation and refinement while separately accounting for offline and per-query costs.

Finally, query-time refinement relies on model-generated interpretations of visual and transcript evidence, whose errors may propagate into refined memory and subsequent decisions. Our prompts encourage uncertainty reporting and evidence-supported updates, while Super-level memory records visual and transcript-derived observations separately. These safeguards do not guarantee correct interpretation or fully resolve conflicts between stored and newly observed information. Future work could track evidence provenance across levels and selectively re-verify uncertain or conflicting updates.

\end{document}